\documentclass[letterpaper]{article} \usepackage[]{aaai2027}  \usepackage[hyphens]{url}  \usepackage{graphicx}    \usepackage{natbib}  \usepackage{caption} \usepackage{amsmath}
\usepackage{amssymb}
\usepackage{booktabs}
\usepackage{fvextra}
\usepackage{algorithm}
\usepackage{algorithmic}

\usepackage{newfloat}
\usepackage{listings}
\DeclareCaptionStyle{ruled}{labelfont=normalfont,labelsep=colon,strut=off} \floatstyle{ruled}
\newfloat{listing}{tb}{lst}{}
\floatname{listing}{Listing}

\graphicspath{{images/PDF/}{images/}{images/png/}}

\title{PlanCraft: Sketch, Refine, and Furnish for Architect-Inspired Progressive 3D Residential Scene Generation}
\author{
    Pengyu Zeng\textsuperscript{*},
    Yuqin Dai\textsuperscript{*},
    Jun Yin\textsuperscript{*},
    Ng Cheuk Hei,
    Ziyang Han,
    Jing Zhong,
    Chaoyang Shi,
    ZhanXiang Jin,
    Maowei Jiang,
        Shuai Lu\textsuperscript{\dag}}
\affiliations{
    Tsinghua University
}

\begin{document}

\maketitle
\begingroup
\renewcommand\thefootnote{}
\footnotetext{$^*$Equal contribution. \\ $^\dag$Corresponding author. E-mail: shuai.lu@sz.tsinghua.edu.cn}
\endgroup
\addtocounter{footnote}{-1}

\begin{abstract}
Two structural insights have been overlooked in automated residential floor plan generation. First, design is inherently progressive. Architects begin with rough strokes and refine them over time, whereas existing methods typically require their conditioning representation to be fully specified before generation, a fundamental mismatch with how design actually works. Second, the 2D floor plan is not an optional intermediate but an irreplaceable spatial contract. Once room boundaries, doors, and windows are fixed, furnishing reduces from open-ended spatial reasoning to bounded constraint satisfaction. Bypassing this contract, as existing 3D systems do by delegating layout to language models, yields overlapping rooms and implausible proportions; directly calling general-purpose language models likewise produces geometrically invalid layouts. Guided by these insights, we present \textbf{PlanCraft}. \textbf{SketchPlan} supplies the missing training signal by replaying the architect's drawing process on 80K real floor plans, producing partial sketches at every completeness level. \textbf{PlanCraft-Diff} progressively sharpens an incomplete sketch into a geometrically precise, vectorizable floor plan through a coarse-to-fine strategy. With the spatial contract established, \textbf{PlanCraft-Agent} then furnishes the scene within well-defined room boundaries. Experiments show that PlanCraft achieves a 61.1\% lower FID than the best existing 2D method and surpasses existing 3D systems by 15 points in expert-rated spatial rationality, with a sketch at only 25\% completion already outperforming all fully specified baselines.
\end{abstract}

\section{Introduction}

\begin{figure}[t]
\centering
\includegraphics[width=\linewidth]{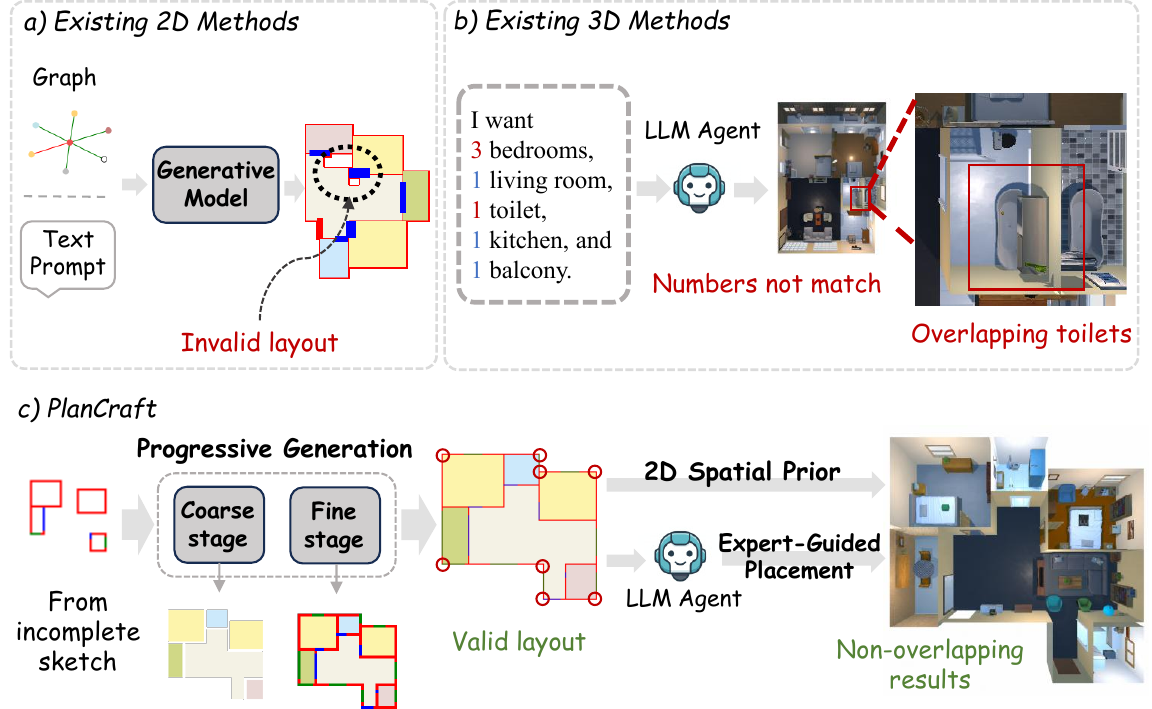}
\caption{\textbf{(a)} Existing 2D methods take a graph or text prompt as input and use generative models to produce a 2D floor plan; they require a fully specified input and may still output an invalid layout (black circle). \textbf{(b)} Existing 3D methods take a text prompt and use an LLM agent to generate a 3D scene directly, without a dedicated data-driven floor-plan generation stage; this can produce incorrect room counts and overlapping objects (red box). \textbf{(c)} PlanCraft takes an incomplete sketch as input. A coarse-to-fine progressive generation stage produces a valid 2D floor plan, which serves as a \textbf{2D spatial prior}. An LLM agent then performs \textbf{expert-guided placement} within the verified room boundaries, yielding non-overlapping 3D results.}
\label{fig:mainfig}
\end{figure}

The design of residential living spaces follows an inherently iterative workflow~\cite{sacks2018bim}. Architects typically begin with incomplete sketches or partial geometric intentions and gradually refine them into explicit floor plans that specify room types, walls, doors, and windows~\cite{hu2020graph2plan, wu2019data}. These floor plans then support downstream 3D visualization for client presentation and design evaluation~\cite{yigitbas2023bimvr}. This pipeline lies at the heart of architectural practice~\cite{sacks2018bim}, yet remains almost entirely manual and time-intensive~\cite{chaillou2020archigan}.

Recent advances in generative models have brought progress to individual stages of this workflow. Floor plan generation methods, including adversarial approaches~\cite{nauata2020house, nauata2021house}, denoising-based methods~\cite{shabani2023housediffusion}, and language model-based approaches~\cite{yin2025floorplan, leng2023tell2design}, can produce 2D layouts from graph or text inputs. Separately, 3D indoor scene generation systems such as Holodeck~\cite{yang2024holodeck} use language model agents to populate virtual environments with furniture. However, these two lines of work remain fundamentally disconnected, as 2D methods stop at floor plans while 3D systems generate their own floor plans internally and do so poorly. Moreover, existing 2D methods typically require their conditioning representation to be fully specified upfront, whether a complete adjacency graph or a detailed text description, a fundamental mismatch with how design actually works.

However, we argue that two structural insights have been overlooked (Fig.~\ref{fig:mainfig}). \textit{Design is progressive}. An architect begins with rough strokes and sharpens them over time, yet every existing method demands a fully specified input upfront. \textit{The floor plan is a spatial prior}. Once room boundaries, doors, and windows are fixed, furnishing reduces from open-ended spatial reasoning to bounded constraint satisfaction. Bypassing this prior, as existing 3D systems do by delegating layout to language models, yields pathological configurations such as a bedroom reachable only by passing through a bathroom, a kitchen twice the size of the living room, or rooms with no accessible entrance. These layouts would be unsellable in practice; directly calling mainstream commercial LLMs~\cite{openai2026gpt55, alibabacloud2026qwen37plus, googledeepmind2025gemini25pro, bytedanceseed2026seed20} likewise produces geometrically invalid layouts.

To address the fact that \textit{design is progressive}, \textbf{SketchPlan} supplies the missing training signal by replaying the architect's drawing process on 80K real floor plans, producing partial sketches at every completeness level. Building on this, \textbf{PlanCraft-Diff} progressively sharpens an incomplete sketch into a geometrically precise, vectorizable floor plan through a coarse-to-fine strategy. To address the fact that \textit{floor plans act as spatial priors}, \textbf{PlanCraft-Agent} furnishes the scene within these verified room boundaries via expert-guided placement. Specifically, a language model generates symbolic spatial constraints from interior design conventions. These constraints are then expanded into geometric form and resolved by a collision-aware solver, replacing unconstrained spatial hallucination with verifiable layout compliance. Experiments show that PlanCraft reduces FID by 2.6--11$\times$ against representative 2D baselines and surpasses Holodeck~\cite{yang2024holodeck} by 15 points in expert-rated spatial rationality, with a sketch at only 25\% completion already outperforming all fully specified baselines. Ablations further show that the full PlanCraft-Diff configuration reduces FID by 3.4$\times$ relative to the one-stage DDPM baseline.

We make the following contributions.
\begin{itemize}
    \item We identify two overlooked structural insights, \emph{design is progressive} and \emph{the floor plan is a spatial prior}, and propose \textbf{PlanCraft}, the first unified system that bridges incomplete design sketches to fully furnished 3D residential scenes.
    \item We introduce \textbf{SketchPlan}, an automated data pipeline that produces partial-sketch training pairs at every completeness level by replaying the architect's drawing process, and \textbf{PlanCraft-Diff}, a unified image-domain diffusion model with two-stage coarse-to-fine training and connectivity-based candidate filtering, which together reduce FID by 2.6--11$\times$ against 2D baselines.
    \item We design \textbf{PlanCraft-Agent}, a grounded 3D assembly stage that combines a language model for constraint generation with rule-based expansion and a collision-aware solver for placement, anchored to verified room boundaries, and show through expert evaluation that it surpasses the prior language model-based 3D system~\cite{yang2024holodeck} in rationality, practicality, and personalization.
\end{itemize}

\section{Related Work}

\subsection{Residential Floor Plan Generation}

Automated floor plan generation has progressed through several paradigms. GAN-based methods treat rooms as spatial units arranged from input topological graphs~\cite{nauata2020house, nauata2021house}. Diffusion-based methods formulate generation as denoising over polygon coordinates~\cite{shabani2023housediffusion} or image-domain pixels~\cite{zeng2024residential, zeng2025rfpa}. Language model-based methods generate layouts from natural language~\cite{leng2023tell2design, yin2025floorplan}. However, all existing paradigms \textbf{\textit{require a fully specified input upfront}}, whether a complete adjacency graph, a bubble diagram, or a detailed text description, and all \textbf{\textit{stop at 2D}}. We instead introduce incomplete design sketches as an input modality that reflects how architects actually work, embodying \textit{design is progressive}, and extend the output to fully furnished 3D scenes.

\subsection{3D Indoor Scene Generation}

3D indoor scene generation targets embodied AI, interior design, and virtual reality. Most prior work arranges furniture within a single predefined room~\cite{feng2023layoutgpt, yang2025llplace, celen2025idesign, lin2024instructscene}, \textbf{\textit{assuming the floor plan is already given}}. Generating complete multi-room scenes is far less explored. ProcTHOR~\cite{deitke2022procthor} relies on procedural rules with \textbf{\textit{limited diversity}}, while Holodeck~\cite{yang2024holodeck} uses language model agents to generate scenes from text but \textbf{\textit{bypasses trained floor plan generation}}, producing \textbf{\textit{spatially incoherent layouts}}. More recent systems have extended language-guided generation toward scalable agentic scenes, multi-floor environments, and traversable interactive worlds~\cite{sage2026,mansion2026,worldgen2026}. These studies broaden the scope of text-conditioned 3D environment generation, but they do not explicitly study progressive completion from incomplete residential sketches. PlanCraft instead focuses on coupling progressive floor-plan completion with geometrically grounded furnishing.

\section{Method}

\begin{figure*}[t]
\centering
\includegraphics[width=\textwidth]{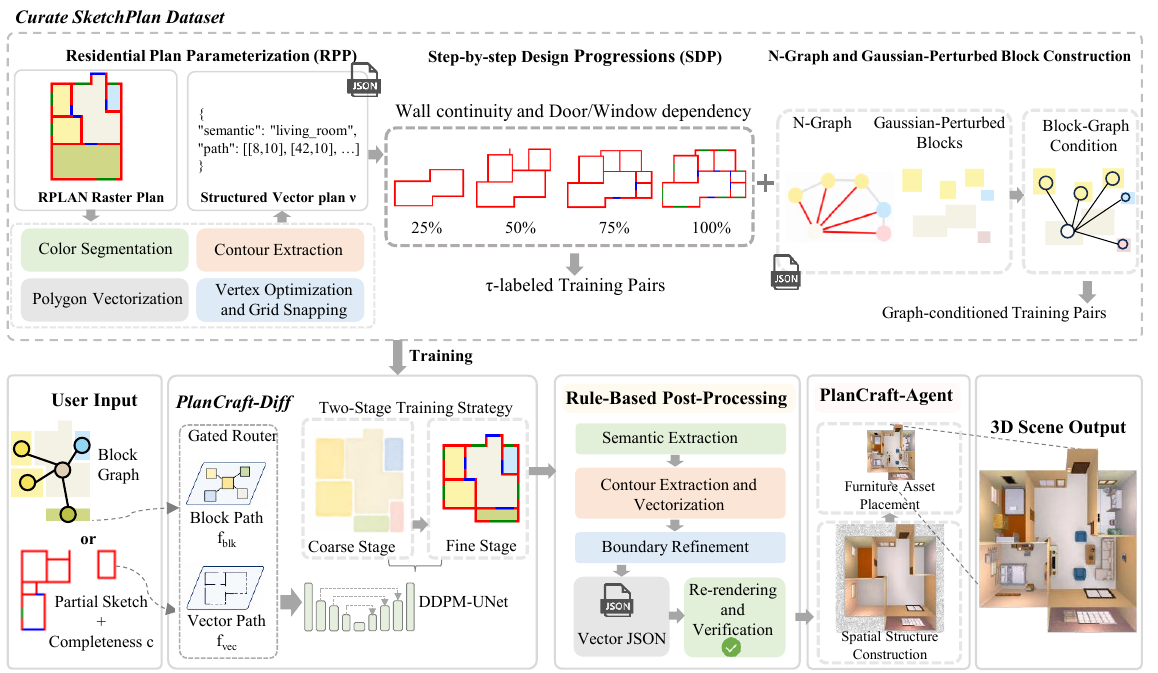}
\caption{Overview of PlanCraft. During training, \textbf{\textit{SketchPlan}} constructs partial-sketch and graph-plus-block supervision for \textbf{\textit{PlanCraft-Diff}}. At inference, a natural-language program is first converted by an LLM into a coarse graph-plus-block condition, while a partial vector sketch is directly rasterized as an alternative condition. PlanCraft-Diff then generates a complete floor plan, which is vectorized and furnished by \textbf{\textit{PlanCraft-Agent}}.}
\label{fig:framework}
\end{figure*}

\subsection{Overview}

PlanCraft is a progressive pipeline (Fig.~\ref{fig:framework}) with four stages. \textbf{SketchPlan} supplies partial-sketch supervision, \textbf{PlanCraft-Diff} refines incomplete inputs into floor plans, \textbf{Rule-Based Post-Processing} converts raster outputs into verified vectors, and \textbf{PlanCraft-Agent} furnishes 3D scenes within verified room boundaries.
Natural-language programs are first converted by an off-the-shelf LLM into coarse graph-plus-block conditions before entering PlanCraft-Diff. PlanCraft supports two input modalities encoded as $H \times H$ multi-channel tensors. A \emph{graph-plus-block input} concatenates a room adjacency graph with per-room-type block maps~\cite{shabani2023housediffusion, nauata2021house}. A \emph{partial vector input} rasterizes the user's incomplete sketch and pads it to the same tensor shape; the completeness ratio $\tau \in [0,1]$ is discretized as $c = \lfloor \tau \times 100 \rfloor$ and used as the conditioning signal for progressive refinement.

\subsection{SketchPlan Dataset}
\label{sec:dataset}

To supply the missing partial-sketch supervision, we construct SketchPlan from RPLAN~\cite{wu2019data}. Three automated pipelines produce paired inputs (graph-plus-block inputs or incomplete vectors) and complete floor plans for training PlanCraft-Diff.

\subsubsection{Residential Plan Parameterization (RPP).}

Each rasterized RPLAN floor plan is converted into a structured vector representation $\mathcal{V} = \{(p_i, s_i)\}_{i=1}^{N}$, where $p_i$ is a polygon path and $s_i \in \mathcal{S}$ its semantic label (room type, door, or window). Color-based segmentation extracts per-category masks, contour detection yields polygon paths, and a vertex optimization step removes collinear points and snaps coordinates to a shared grid. The vectorized plans serve as the source for the two pipelines below.

\subsubsection{Step-by-step Design Progressions (SDP).}

From each $\mathcal{V}$, we generate a sequence $\mathcal{V}^{(0)} \subset \mathcal{V}^{(1)} \subset \cdots \subset \mathcal{V}^{(K)} = \mathcal{V}$ that simulates progressive drafting. An element $v_i$ is eligible at step $k$ only if its prerequisite walls already appear in $\mathcal{V}^{(k)}$; we then sample eligible walls with probability $\lambda$ and doors/windows with $1-\lambda$, yielding training pairs $(\mathcal{V}^{(k)}, \mathcal{V})$ at diverse completeness levels $\tau = |\mathcal{V}^{(k)}|/|\mathcal{V}|$. Randomized drawing orders and $\lambda$ values provide multiple sequences per plan.

\subsubsection{N-Graph and Gaussian-Perturbed Block Construction.}

For graph-conditioned inputs, we build two complementary representations. The \textbf{N-graph} enriches the conventional attributed graph with room position and wall connectivity, where each room is a node at position $\mathbf{p}_i \sim \mathcal{U}(\text{bbox}_i)$, with edges encoding wall-sharing adjacency. \textbf{Gaussian-perturbed room blocks} extract each room-type spatial extent as a block image whose center is displaced by $\boldsymbol{\delta} \sim \mathcal{N}(0, \sigma_\text{block}^2 \mathbf{I}_2)$. The N-graph and room-type block maps are concatenated into the graph-plus-block conditional input consumed by PlanCraft-Diff.

\subsection{PlanCraft-Diff: Progressive Floor Plan Refinement}
\label{sec:diff}

Armed with SketchPlan's training pairs, PlanCraft-Diff uses a conditional denoising diffusion probabilistic model (DDPM)~\cite{ho2020denoising} with a coarse-to-fine strategy to balance global plausibility and geometric precision.

\textbf{Forward Process.} Given a clean floor plan image $\mathbf{x}_0$, the forward process produces a noisy version at timestep $t$,
\begin{equation}
    \mathbf{x}_t = \sqrt{\bar{\alpha}_t}\,\mathbf{x}_0 + \sqrt{1-\bar{\alpha}_t}\,\boldsymbol{\epsilon},
\end{equation}
where $\boldsymbol{\epsilon} \sim \mathcal{N}(0, \mathbf{I})$, $\alpha_t = 1 - \beta_t$, and $\bar{\alpha}_t = \prod_{s=1}^{t}\alpha_s$; the variance schedule $\{\beta_t\}_{t=1}^{T}$ increases linearly from $\beta_1$ to $\beta_T$.

\textbf{Conditional Reverse Process.} A noise prediction network $\epsilon_\theta$ is conditioned on the noisy image $\mathbf{x}_t$, timestep $t$, the conditional input $\mathbf{x}_\text{cond}$, and the completeness signal $c$. The backbone is a UNet~\cite{ronneberger2015u} with four resolution levels at base width $C$. Because graph-plus-block inputs and partial vectors encode different structures, we use modality-specific entry convolutions,
\begin{equation}
    \mathbf{h}_0 = \begin{cases} f_\text{vec}(\mathbf{x}_t, \mathbf{x}_\text{cond}) & \text{if } c \text{ is present,} \\ f_\text{blk}(\mathbf{x}_t, \mathbf{x}_\text{cond}) & \text{otherwise,} \end{cases}
\end{equation}
where $f_\text{vec}$ and $f_\text{blk}$ both map the conditional tensor to width $C$. The timestep $t$ is encoded via sinusoidal positional embeddings~\cite{vaswani2017attention} and injected into each residual block as an additive bias, and $c$ is encoded with a learnable embedding table and injected in the same way. This design lets one unified model handle both modalities without collapsing them into the same low-level representation, while still sharing most parameters in deeper layers.

\textbf{Training Objective.} The model minimizes the standard $\ell_2$ noise prediction loss,
\begin{equation}
    \mathcal{L} = \mathbb{E}_{\mathbf{x}_0, \boldsymbol{\epsilon}, t}\left[\|\boldsymbol{\epsilon} - \epsilon_\theta(\mathbf{x}_t, t, \mathbf{x}_\text{cond}, c)\|^2\right].
\end{equation}

\textbf{Two-Stage Training Strategy.} We train in two stages. Stage~1 pretrains for $E_1$ epochs on coarse blurred layouts (using Gaussian perturbation with scale $\sigma_\text{blur}$ over ground-truth room blocks and graphs) to learn global spatial distributions, and Stage~2 fine-tunes for $E_2$ epochs on high-fidelity floor plans to sharpen boundary precision. This strategy reduces FID from 13.810 to 11.674 (15.5\% improvement) over single-stage training (Tab.~\ref{tab:ablation}).

\textbf{Inference and Quality Filtering.} At inference, the model denoises from Gaussian noise over $T$ steps and generates $N_s$ candidates per input. For the natural-language route, an off-the-shelf LLM converts each program into a coarse graph-plus-block condition. Candidates whose floor plan region is not a single connected component (after removing door/window markings) are discarded. This connectivity check reduces FID from 11.674 to 4.026 (Tab.~\ref{tab:ablation}); only verified plans proceed to post-processing.

\subsection{From Image to Vector: Rule-Based Post-Processing}
\label{sec:postproc}

The generated raster floor plan is converted into a structured vector representation for downstream 3D asset placement. We apply color-range thresholding in BGR space to extract per-category masks, detect contours, and convert them into polygon paths. Room contours are dilated by one pixel so that adjacent rooms share boundaries, and coordinates are snapped to a shared grid to enforce wall alignment. The refined paths are saved as JSON and re-rendered for verification; only verified vectors are forwarded to PlanCraft-Agent.

\subsection{PlanCraft-Agent: Grounded 3D Scene Assembly}
\label{sec:agent}

With a verified vector floor plan as a \textbf{2D spatial prior}, \textbf{PlanCraft-Agent} performs expert-guided placement within verified room boundaries, yielding a +15-point advantage in expert-rated Rationality over Holodeck~\cite{yang2024holodeck}.

Given the verified vector floor plan, PlanCraft-Agent assembles the 3D scene in two phases. In \textbf{Spatial Structure Construction}, room polygons are converted from pixel to metric coordinates at resolution $r$ pixels per meter, while specialized agents build wall meshes, place door assets retrieved from Objaverse~\cite{deitke2023objaverse} via CLIP~\cite{radford2021learning}, and determine window types and sizes. In \textbf{Furniture Asset Selection and Placement}, an object-selection agent retrieves room-specific furniture via CLIP~\cite{radford2021learning} and Sentence-BERT~\cite{reimers2019sbert}, a constraint-generation agent emits spatial constraints, and a placement engine solves for collision-free 3D positions and orientations within room boundaries. The final scene is serialized and deployed in AI2-THOR~\cite{kolve2017ai2thor}. Additional details of the agent roles, spatial constraints, and placement procedure are provided in the supplementary material.

\section{Experiments}

\begin{table*}[t]
\centering
\resizebox{0.97\textwidth}{!}{\begin{tabular}{lccccc}
\toprule
\textbf{Method} & \textbf{Venue} & \textbf{FID}$\downarrow$ & \textbf{IoU}$\uparrow$ & \textbf{PSNR}$\uparrow$ & \textbf{SSIM}$\uparrow$ \\
\midrule
\multicolumn{6}{l}{\textit{2D Floor Plan Generation Methods}} \\
\midrule
HouseGAN~\cite{nauata2020house} & ECCV 2020 & 44.45 & 0.220 & 10.264 & 0.621 \\
HouseGAN++~\cite{nauata2021house} & CVPR 2021 & 32.05 & 0.319 & 11.528 & 0.685 \\
HouseDiffusion~\cite{shabani2023housediffusion} & CVPR 2023 & 10.35 & 0.412 & 12.847 & 0.743 \\
Floorplan-LLaMA~\cite{yin2025floorplan} & ACL 2025 & 22.61 & 0.285 & 11.105 & 0.662 \\
Tell2Design~\cite{leng2023tell2design} & ACL 2023 & 11.754 & 0.395 & 12.563 & 0.721 \\
\midrule
\multicolumn{6}{l}{\textit{3D Generation Methods}} \\
\midrule
Holodeck~\cite{yang2024holodeck} & CVPR 2024 & 47.82 & 0.089 & 8.952 & 0.583 \\
\midrule
PlanCraft-Diff (Block Graph) & Ours & \textbf{4.026}~\textcolor{green!60!black}{\scriptsize($\downarrow$61.1\%)} & \textbf{0.510}~\textcolor{green!60!black}{\scriptsize($\uparrow$23.8\%)} & \textbf{13.758}~\textcolor{green!60!black}{\scriptsize($\uparrow$7.1\%)} & \textbf{0.830}~\textcolor{green!60!black}{\scriptsize($\uparrow$11.7\%)} \\
\bottomrule
\end{tabular}}
\caption{Quantitative comparison against 2D floor plan generation methods and 3D generation methods. Holodeck vector floor plans are mapped to RPLAN semantic labels prior to evaluation. PlanCraft-Diff achieves the best reported performance across all metrics in this system-level comparison. $\uparrow$ / $\downarrow$ indicate higher-is-better / lower-is-better. Green annotations show PlanCraft's improvement over the strongest 2D baseline (HouseDiffusion).}
\label{tab:main}
\end{table*}

\subsection{Experimental Setup}

\textbf{Datasets.} For 2D generation, we build training and evaluation data from RPLAN~\cite{wu2019data}, a corpus of over 80K real residential floor plans. To simulate incomplete design sketches, our automated pipeline progressively masks structural elements from complete plans, producing inputs at 25\%, 50\%, 75\%, and 100\% vector completeness. Our system also supports graph-plus-block inputs, enabling a system-level comparison with prior graph- and language-conditioned floor-plan generation methods. For 3D furniture placement, we adopt the same publicly available asset database used by Holodeck~\cite{yang2024holodeck}.

\textbf{Baselines.} We compare against representative GAN-based methods (HouseGAN~\cite{nauata2020house}, HouseGAN++~\cite{nauata2021house}), diffusion-based methods (HouseDiffusion~\cite{shabani2023housediffusion}), language model-based methods (Floorplan-LLaMA~\cite{yin2025floorplan}, Tell2Design~\cite{leng2023tell2design}), and the 3D system Holodeck~\cite{yang2024holodeck}.

\textbf{Evaluation Metrics.} For 2D evaluation, we follow prior work~\cite{zeng2025rfpa, shabani2023housediffusion, nauata2021house} and report FID~\cite{heusel2017gans}, IoU~\cite{rezatofighi2019generalized}, PSNR~\cite{hore2010image}, and SSIM~\cite{wang2004image}. For 3D evaluation, we conduct expert assessment because no automatic metric adequately captures furnished residential scene quality.

\textbf{Implementation Details.} PlanCraft-Diff is a conditional DDPM~\cite{ho2020denoising} with a UNet~\cite{ronneberger2015u} backbone at base width $C=64$, trained on images resized to $H \times H$ with $H=64$. We use a linear noise schedule with $T=1000$ steps, $\beta_1=1\times10^{-4}$, and $\beta_T=0.02$. We use Adam~\cite{kingma2015adam} with initial learning rate $\eta=2\times10^{-4}$ annealed to $\eta_\text{min}=2\times10^{-6}$ via cosine schedule~\cite{loshchilov2017sgdr}, Adam momenta $(0.9,0.999)$, no weight decay, and batch size $B=64$. We pretrain for $E_1{=}1000$ epochs on coarse-grained blurred images and fine-tune for $E_2{=}500$ epochs on high-fidelity images. At inference, $N_s=12$ candidates are generated per input. All experiments run on an NVIDIA A100 (80GB) GPU using PyTorch. For 3D deployment, the scene JSON is loaded into AI2-THOR~\cite{kolve2017ai2thor} (v5.0.0), in which all 3D evaluations are also conducted.

\subsection{Quantitative Comparison on Floor Plan Generation}

Tab.~\ref{tab:main} presents the quantitative comparison of 2D floor plan generation.

\textit{PlanCraft outperforms all baselines across all metrics.} PlanCraft-Diff achieves the best FID, IoU, PSNR, and SSIM. The largest gain appears on FID, indicating more faithful global floor-plan distributions, while the gains on IoU and SSIM show that the improvement also extends to structural coherence and perceptual quality. This pattern is consistent with our design goal, where coarse-to-fine training first captures plausible global spatial organization and then sharpens local geometry.

\textit{Explicit 2D constraints remain necessary for geometric validity.} GAN, diffusion, and language-model baselines all underperform PlanCraft-Diff, and Holodeck~\cite{yang2024holodeck} ranks last, as it bypasses principled 2D floor-plan generation. This result supports our claim that language-model reasoning alone is insufficient for coherent residential layout generation without an explicit 2D constraint stage. The results suggest that combining room-topology information with approximate spatial blocks provides an effective structural condition for residential floor-plan generation.

\subsection{Effect of Vector Completeness}
\label{sec:completeness}

\begin{figure*}[t]
\centering
\includegraphics[width=\textwidth]{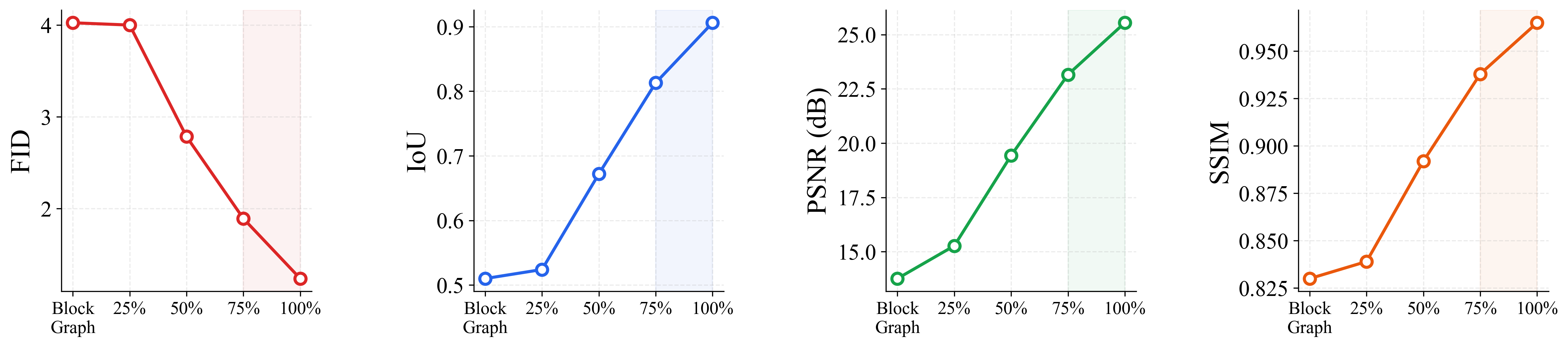}
\caption{Effect of vector completeness on generation quality. All four metrics improve monotonically, but at different rates: PSNR and IoU respond most steeply to additional structural vectors, while SSIM improves more gradually, indicating that perceptual layout quality is already well captured at low completeness. Even at 25\%, PlanCraft-Diff outperforms all fully specified baselines in Tab.~\ref{tab:main}.}
\label{fig:completeness_curve}
\end{figure*}

Beyond the standard graph-plus-block comparison above, PlanCraft's distinctive input modality is partial design sketches at arbitrary completeness. Fig.~\ref{fig:completeness_curve} traces all four metrics as vector completeness increases from 25\% to 100\%, with the graph-plus-block setting included as a zero-vector reference point.

\textit{Additional vectors mainly sharpen geometry.} Fig.~\ref{fig:completeness_curve} shows that PSNR and IoU respond most strongly at 25--50\% completeness, while SSIM changes more gradually. This suggests that sparse sketches already establish coarse room zoning, and later vectors mainly refine wall placement.

Even at 25\% completeness, PlanCraft surpasses all fully specified baselines in Tab.~\ref{tab:main}, showing that sparse sketches already support strong completion. The slower SSIM growth indicates that early vectors establish coarse zoning while later vectors mainly sharpen geometry.

\subsection{Expert Evaluation on 3D Scene Generation}
\label{sec:expert}

Holodeck~\cite{yang2024holodeck} is the closest prior language-guided system that generates prompt-conditioned furnished multi-room scenes. We conduct a user study with 26 participants with architectural training, including 20 master's-level and six doctoral-level participants. Each participant used both PlanCraft and Holodeck to complete 20 prompt-matched residential scene-generation tasks. After interacting with each system, participants rated it on a 100-point scale along three dimensions of Rationality, Practicality, and Personalization. The reported scores are averaged over all tasks and participants. Detailed evaluation criteria are provided in the supplementary material.

\begin{figure}[t]
\centering
\includegraphics[width=0.85\columnwidth]{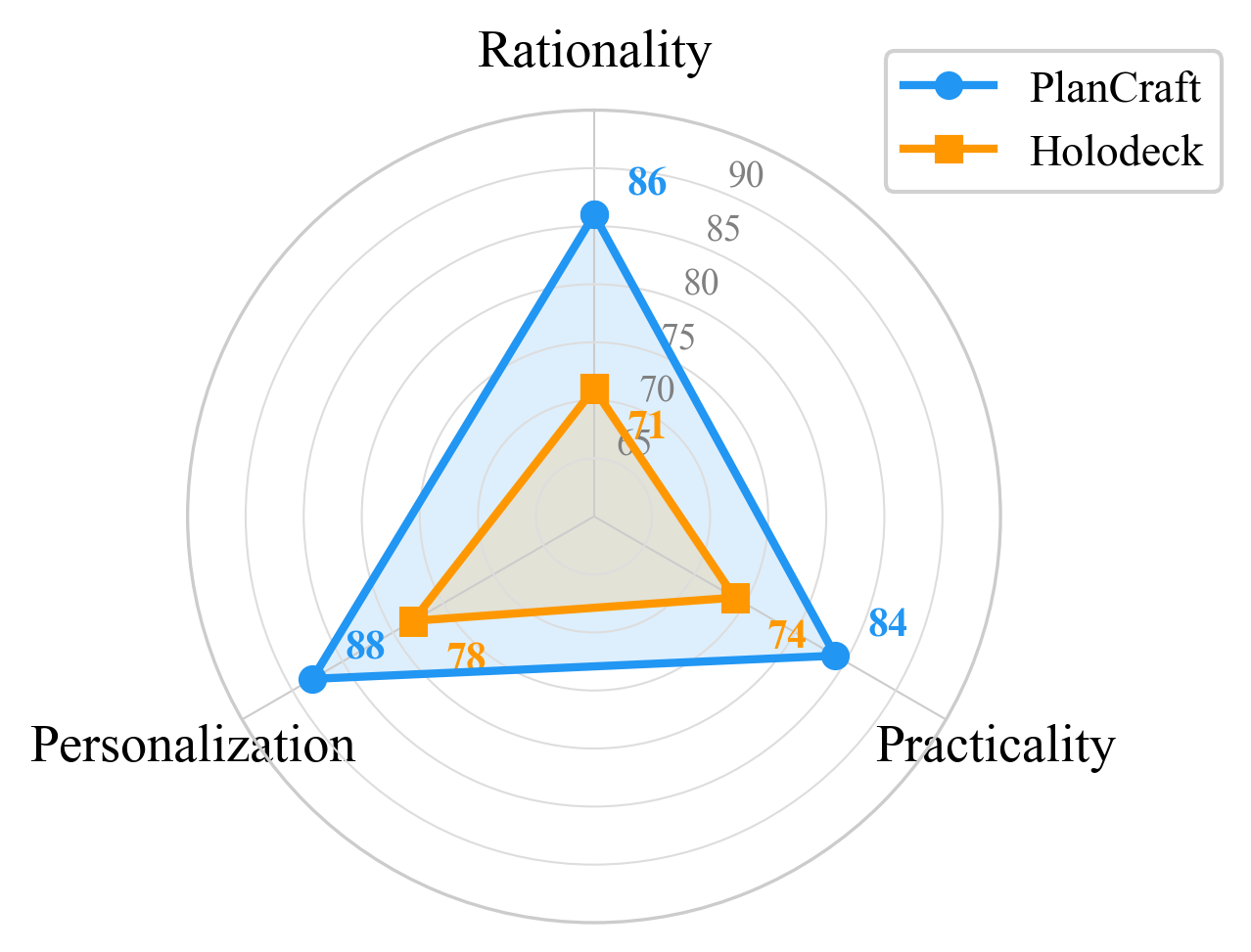}
\caption{Expert evaluation on 3D scene generation. Scores are on a 100-point scale across three dimensions. PlanCraft outperforms Holodeck~\cite{yang2024holodeck} across all dimensions, with the largest gap in Rationality (+15).}
\label{fig:expert}
\end{figure}

As shown in Fig.~\ref{fig:expert}, PlanCraft outperforms Holodeck~\cite{yang2024holodeck} across all three dimensions, with the largest gap in Rationality. The gap arises because PlanCraft establishes verified room boundaries before furniture placement, whereas Holodeck relies on unconstrained language-model reasoning. The Personalization advantage indicates stronger alignment with the prompt-specific spatial and furnishing requirements; the Practicality gap is smaller because both systems accept natural language and produce deployable 3D scenes. Importantly, the expert study shows that improvements in 2D floor-plan validity propagate to the user-visible 3D scene rather than remaining confined to intermediate representations.

\subsection{Qualitative Comparison}

\begin{figure}[t]
\centering
\includegraphics[width=\columnwidth]{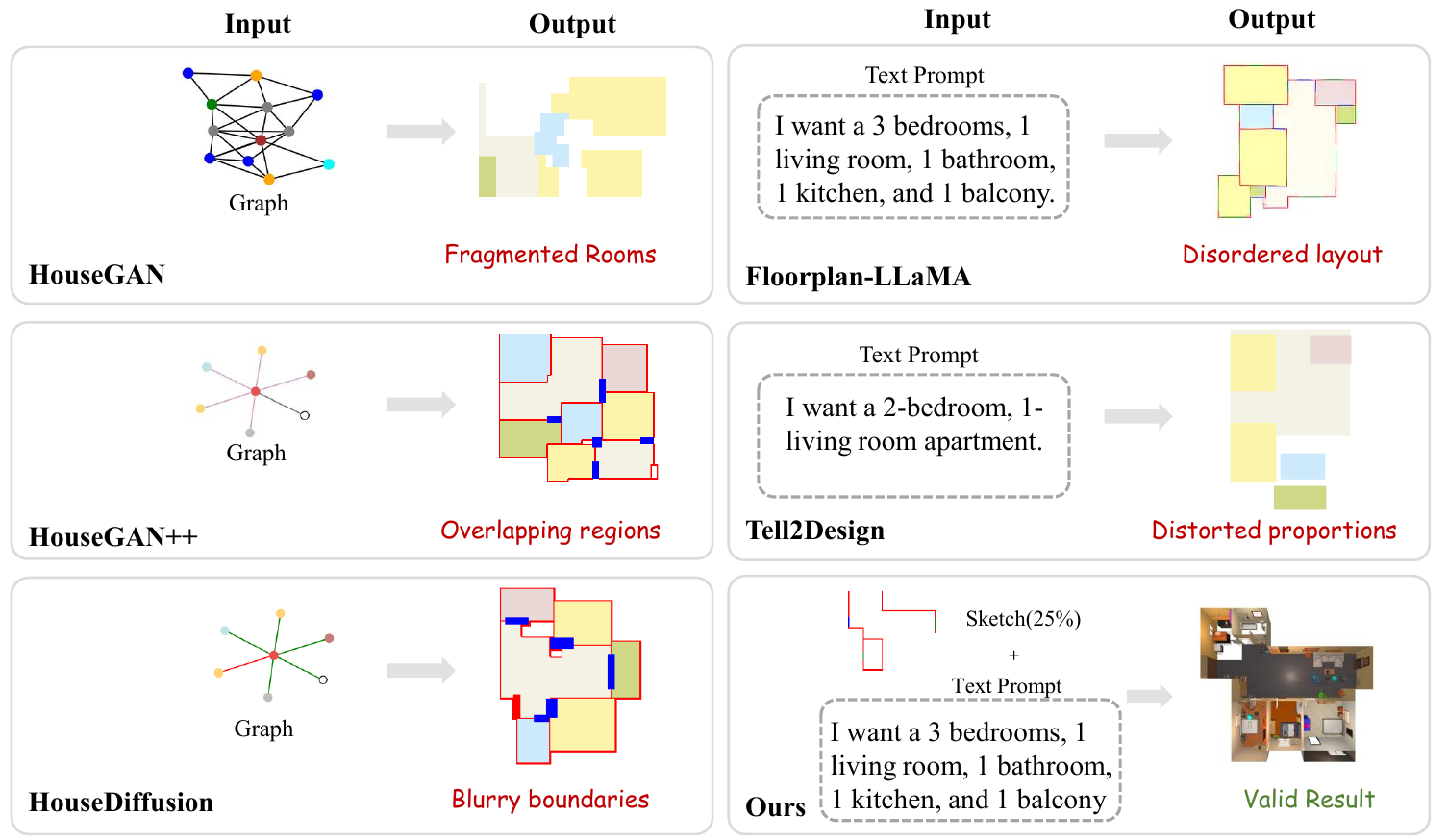}
\caption{Representative outputs and failure cases across floor plan generation methods. Each panel shows a method's input (left) and output (right), with baseline failure modes annotated in red. HouseGAN and HouseGAN++ take bubble diagrams, HouseDiffusion takes room connectivity graphs, and Floorplan-LLaMA and Tell2Design accept natural language. PlanCraft supports language-guided editing of incomplete vector sketches: an LLM first updates the sketch according to the text instruction, and the resulting vector representation is then rasterized as the input to PlanCraft-Diff.}
\label{fig:qualitative_methods}
\end{figure}

\begin{figure*}[t]
\centering
\includegraphics[width=\textwidth]{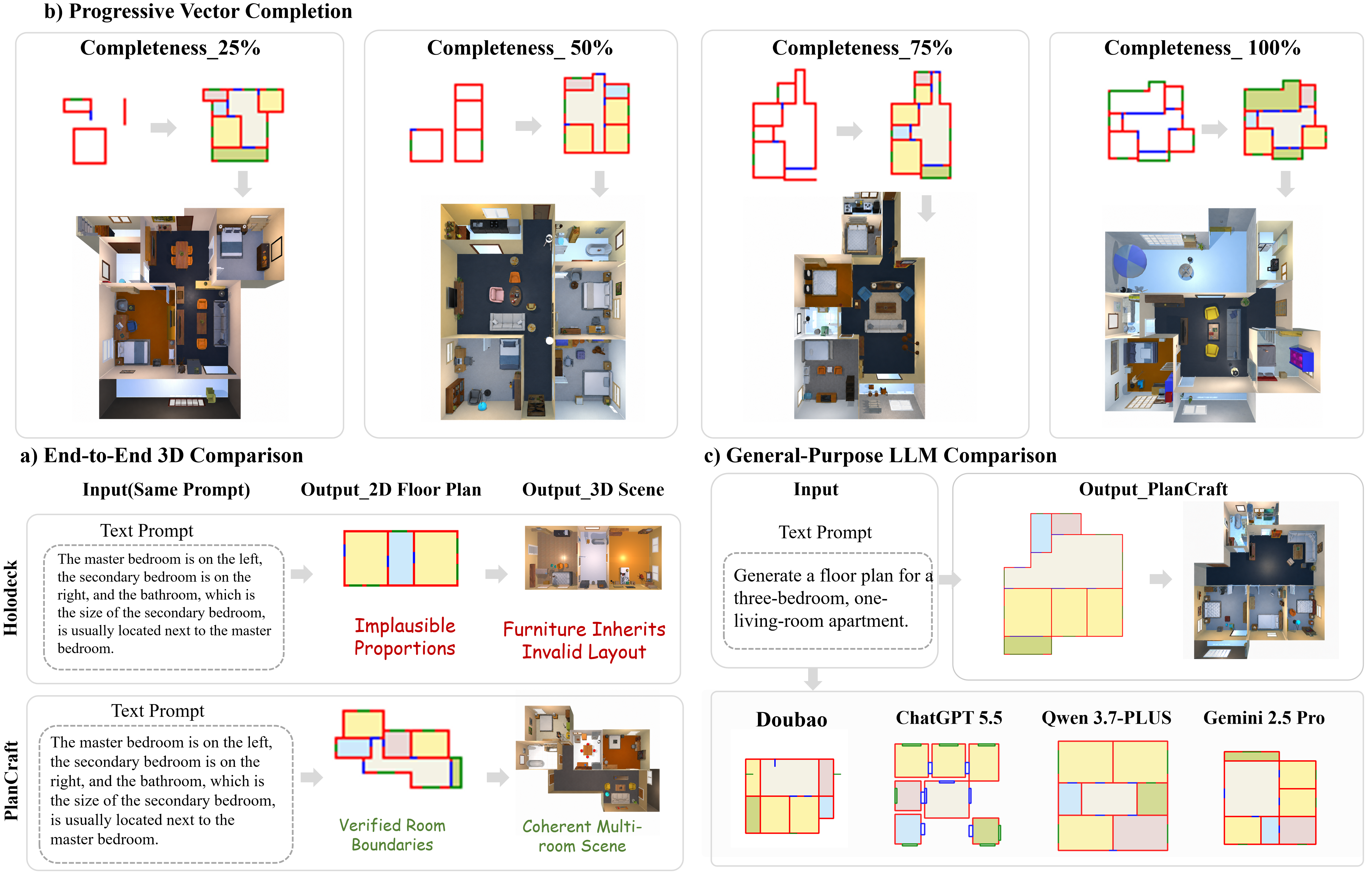}
\caption{Qualitative comparison from three perspectives. \textbf{(a)} End-to-end comparison with Holodeck~\cite{yang2024holodeck}. Given the same natural language description, PlanCraft generates spatially coherent floor plans that translate into realistic 3D scenes, while Holodeck suffers from invalid geometry: bedrooms accessible only via bathrooms, and disproportionate spaces (kitchens as the largest room, exceeding living rooms). \textbf{(b)} PlanCraft under varying vector completeness (25\%, 50\%, 75\%, 100\%). Even with only 25\% of structural vectors provided, PlanCraft infers a plausible complete layout. \textbf{(c)} Comparison with language models. All four models produce layouts with significant deficiencies (annotated in red), while PlanCraft generates well-organized layouts.}
\label{fig:qualitative_combined}
\end{figure*}

\textbf{Comparison across floor plan generation methods (Fig.~\ref{fig:qualitative_methods}).} Baselines exhibit paradigm-specific failure modes, including fragmented rooms, distorted proportions, or geometrically invalid layouts. PlanCraft avoids these failures by combining image-domain generation with a coarse-to-fine strategy and an explicit 2D constraint stage. The comparison also highlights that architectural plausibility depends on enforcing both room topology and geometric closure rather than optimizing text-image consistency alone.

\textbf{Qualitative evidence across 3D generation, sketch completeness, and language models (Fig.~\ref{fig:qualitative_combined}).} The three comparisons converge on the same bottleneck. Without an explicit 2D spatial prior, both Holodeck and standalone language models (Doubao~\cite{bytedanceseed2026seed20}, ChatGPT~\cite{openai2026gpt55}, Qwen~\cite{alibabacloud2026qwen37plus}, Gemini~\cite{googledeepmind2025gemini25pro}) produce layouts with invalid geometry, while PlanCraft preserves coherent room structure from sparse sketches to complete reconstructions.

Fig.~\ref{fig:qualitative_combined}(c) shows that Doubao, ChatGPT, Qwen, and Gemini all produce invalid geometry, whereas PlanCraft maintains coherent room structure after grounding the request in a 2D sketch prior. This contrast confirms that 2D spatial priors are necessary for high-quality 3D scene generation.

\subsection{Editability}

PlanCraft supports two editing modalities. In \textbf{Natural Language Editing}, the generated floor plan is abstracted into a graph-plus-block representation, edited from natural language instructions, then regenerated and refurnished. For partial-sketch editing, the LLM directly updates the vector primitives; for editing a generated plan, it operates on the abstracted graph-plus-block representation. In \textbf{CAD Vector Editing}, the plan is exported as CAD-compatible vectors, manually edited, and fed back as updated conditional inputs.

\subsection{Ablation Study}

\begin{table}[t]
\centering
\resizebox{\columnwidth}{!}{\begin{tabular}{lcccc}
\toprule
\textbf{Variant} & \textbf{FID}$\downarrow$ & \textbf{IoU}$\uparrow$ & \textbf{PSNR}$\uparrow$ & \textbf{SSIM}$\uparrow$ \\
\midrule
MeanFlow backbone        & 30.643 & 0.318 & 10.892 & 0.654 \\
DDPM (one-stage)         & 13.810 & 0.423 & 12.106 & 0.742 \\
DDPM (no filtering)     & 11.674 & 0.498 & 13.521 & 0.815 \\
\textbf{PlanCraft-Diff (full)} & \textbf{4.026}~\textcolor{green!60!black}{\scriptsize($\downarrow$65.5\%)} & \textbf{0.510}~\textcolor{green!60!black}{\scriptsize($\uparrow$2.4\%)} & \textbf{13.758}~\textcolor{green!60!black}{\scriptsize($\uparrow$1.8\%)} & \textbf{0.830}~\textcolor{green!60!black}{\scriptsize($\uparrow$1.8\%)} \\
\bottomrule
\end{tabular}}
\caption{Ablation study under the graph-plus-block input setting. We validate each key design choice of PlanCraft-Diff. The full model (bold) consistently achieves the best performance across all four metrics.}
\label{tab:ablation}
\end{table}

We conduct ablation experiments to validate our key design choices. Tab.~\ref{tab:ablation} summarizes the main results; additional analyses are provided in the supplementary material.

\textbf{Generative Architecture: DDPM vs.\ MeanFlow.} DDPM dramatically outperforms MeanFlow~\cite{geng2025meanflow} under matched training conditions (Tab.~\ref{tab:ablation}). We attribute the gap to the structured, discrete nature of floor plans, where iterative denoising better preserves aligned walls and closed polygons. By contrast, direct transport in flow matching is less robust when tiny local errors can break topological validity.

\textbf{Training Strategy: Two-Stage vs.\ One-Stage.} Removing coarse pretraining substantially degrades FID while affecting IoU, PSNR, and SSIM less. This indicates that two-stage training primarily improves population-level distribution learning before fine-grained geometric refinement. The result is important because it shows that better floor-plan generation requires not only a stronger backbone but also a curriculum aligned with progressive design.

\textbf{Effect of Candidate Filtering.} Candidate filtering sharply improves FID while leaving other metrics nearly unchanged, because it removes disconnected outliers rather than altering the spatial layout of already valid samples. This confirms that the candidate-filtering stage serves mainly as a reliability filter instead of a hidden source of geometric improvement.

\section{Conclusion}

We present PlanCraft, a unified system that bridges incomplete design sketches to fully furnished 3D residential scenes. Grounded in two structural insights, \textbf{design is progressive} and \textbf{the floor plan is a spatial prior}, PlanCraft integrates SketchPlan, PlanCraft-Diff, rule-based vector post-processing, and PlanCraft-Agent into a unified workflow. SketchPlan supplies partial-sketch training data, PlanCraft-Diff sharpens an incomplete sketch into a geometrically valid floor plan via coarse-to-fine diffusion, and PlanCraft-Agent performs expert-guided placement within verified room boundaries. PlanCraft reduces FID over 2D baselines and surpasses the prior 3D system in expert-rated rationality.

\end{document}